\documentclass{article}
\usepackage[final]{neurips_2021}
\makeatletter
\renewcommand{\@noticestring}{%
1st Workshop on Human and Machine Decisions
(WHMD 2021) at NeurIPS 2021.}
\makeatother

\usepackage[utf8]{inputenc} 
\usepackage[T1]{fontenc}    
\usepackage{hyperref}       
\usepackage{url}            
\usepackage{booktabs}       
\usepackage{amsfonts}       
\usepackage{nicefrac}       
\usepackage{microtype}      
\usepackage{xcolor}         
\usepackage{amsmath}
\usepackage{times}  
\usepackage{helvet} 
\usepackage{courier}  
\usepackage{graphicx}
\usepackage{amsmath}
\usepackage{natbib}  
\usepackage{caption} 

\usepackage{setspace}
\title{Improving Learning-to-Defer Algorithms Through Fine-Tuning}

\author{%
  Naveen Raman \\
  Department of Computer Science\\
  University of Maryland\\
  College Park, MD \\
  \texttt{nav.j.raman@gmail.com} \\
  \and
  \textbf{Michael Yee} \\
  MIT Lincoln Laboratory\\
  Lexington, MA\\
  \texttt{myee@ll.mit.edu} \\
}

\begin{document}

\maketitle

\makeatletter{\renewcommand*{\@makefnmark}{}
\footnotetext{DISTRIBUTION STATEMENT A. Approved for public release. Distribution is unlimited. This material is based upon work supported by the Under Secretary of Defense for Research and Engineering under Air Force Contract No. FA8702-15-D-0001. Any opinions, findings, conclusions or recommendations expressed in this material are those of the author(s) and do not necessarily reflect the views of the Under Secretary of Defense for Research and Engineering. © 2021 Massachusetts Institute of Technology. Delivered to the U.S. Government with Unlimited Rights, as defined in DFARS Part 252.227-7013 or 7014 (Feb 2014). Notwithstanding any copyright notice, U.S. Government rights in this work are defined by DFARS 252.227-7013 or DFARS 252.227-7014 as detailed above. Use of this work other than as specifically authorized by the U.S. Government may violate any copyrights that exist in this work.}\makeatother}

\begin{abstract}
The ubiquity of AI leads to situations where humans and AI work together, creating the need for learning-to-defer algorithms that determine how to partition tasks between AI and humans. 
We work to improve learning-to-defer algorithms when paired with specific individuals by incorporating two fine-tuning algorithms and testing their efficacy using both synthetic and image datasets. 
We find that fine-tuning can pick up on simple human skill patterns, but struggles with nuance, and we suggest future work that uses robust semi-supervised to improve learning.  
\end{abstract}

\section{Introduction}
The rise of neural technologies has led to the adoption of AI-based solutions for human-facing tasks in fields such as health care~[\cite{tran2019patients}] and criminal justice~\cite{washington2018argue}.
In some situations, humans and AI collaborate to solve a problem by partitioning tasks so AI solves problems that are difficult for humans. 
Determining task partitions is done through a learning-to-defer algorithm, which uses a human performance model to take human skill into account. 
Numerous studies~[\cite{mozannar2020consistent,raghu2019algorithmic,wilder2020learning}] have developed learning-to-defer algorithms using confidence-based~[\cite{raghu2019algorithmic} and joint learning methods~[\cite{mozannar2020consistent,okati2021differentiable}], which learn rejector and classifier functions simultaneously. 

Learning-to-defer algorithms can be used with specific individuals, such as an individual doctor working with an AI to classify X-ray images. 
In these situations, learning-to-defer algorithms can be improved by fine-tuning their underlying human performance models. 
Fine-tuning exploits human skill variation to improve system performance. 
For example, when AI classifies X-ray images, learning-to-defer algorithms should tune to the doctor's expertise. 
We incorporate fine-tuning into learning-to-defer algorithms, addressing an open research question ~[\cite{de2021classification}]. 
To do this, we develop two fine-tuning methods and demonstrate their efficacy on both synthetic and image data~[\cite{krizhevsky2009learning}].
Our contributions are as follows. 
We (1) develop fine-tuning methods which allow human-AI models to be individual specific and (2) apply this to autonomous driving and image recognition tasks.
    
\section{Problem Statement}
We define the problem of human-AI collaboration for classification and  learning-to-defer algorithms.
We consider the problem of classifying data points $x_{1}, \ldots, x_{n}$, which have true labels $y_{1}, \ldots, y_{n}$. 
Data points $x_{1}, \ldots, x_{n}$ have human annotations $h_{1}, \ldots, h_{n}$. 
The problem is to develop functions $a$ and $m$ that partition and solve the task, so if $a(x_i) = 0$, the task is solved by a human, and if $a(x_i) = 1$, then the task is solved by a machine, $m$, which is a classification function. 
The objective function is then to maximize 
\begin{equation} 
\sum_{i=0}^{n}  
\begin{cases} 
      h_i=y_i & a(x_i) = 0 \\
      m(x_i)=y_i & a(x_i) = 1
   \end{cases}. 
\end{equation}
To develop the function $a$, we're given the following: 
\begin{enumerate}
    \item Aggregate human data: a set of points (for training) consisting of data points $w_1, \ldots, w_m$, human labels $g_1, \ldots, g_m$ (aggregated across multiple humans), and true labels $z_1, \ldots, z_m$. 
    \item Specific human data: a set of points (for fine-tuning and testing) consisting of data points $v_1, \ldots, v_k$ ($k < m$), human labels from an individual $f_1, \ldots, f_k$, and true labels $t_1, \ldots, t_m$.
    \item Non-human labeled data points: a set of points (for imputing human labels) consisting of data points $u_1, \ldots, u_p$ ($k < m < p$) and truth labels $s_1, \ldots, s_p$.
\end{enumerate}

\section{Learning-to-defer algorithms}
We describe three learning-to-defer algorithms, each of which uses data in different ways to develop the function $a$: a non-fine-tuning baseline and two approaches that incorporate fine-tuning. 
\label{sec: methods}
\subsection{Baseline Non-fine-tuning}
We describe the state-of-the-art learning-to-defer algorithm~[\cite{mozannar2020consistent}]. 
The algorithm first develops a human performance model, $\hat{h}$, which is trained on all human performance data, $w_i$ and $v_i$, and outputs whether a human will answer a data point correctly.
After learning $\hat{h}$, human performance labels are imputed on data points $u_{1} \cdots u_{p}$, so all three data types have human labels. 
Afterward, the learning-to-defer function $a$, and the machine classification function, $m$, are learned simultaneously by reformulating as a $k+1$ classification problem and using a modified cross-entropy loss function that takes into account the cost of deferral. 
See Section~[\cite{mozannar2020consistent}] for full details of this approach. 


\subsection{Baseline Fine-tuning} 
We incorporate elements of fine-tuning into the state-of-the-art learning-to-defer algorithm to improve system accuracy for specific individuals.
To do this, we modify the human performance model training, so $\hat{h}$ is first trained for $n$ epochs on aggregate human performance data, $w_i$, then trained for $n \lambda$ epochs on specific human performance data, $v_i$. 
To account for class imbalances, which could skew training due to small fine-tuning dataset size, we introduce a class-weighting scheme, which weights the loss function by the inverse frequency of each class. 
\subsection{Self-training} 

We adopt the self-training algorithm from semi-supervised learning as an alternative to the baseline~[\cite{chapelle2009semi}]. 
For self-training, we first train a human performance model using specific human data $v_i$.
Using this basic model, we predict labels for the unlabeled data points, $u_{i}$, then select the high confidence predictions.
We use these high confidence points to retrain the model, thereby using unlabeled data points to develop an individual-specific model.  
\section{Autonomous Vehicle Experiment}
\label{sec:driving}
Human-AI collaboration naturally arises in the realm of autonomous vehicles, which requires deferral to humans depending on environmental circumstances. 
We develop a synthetic dataset based on autonomous driving, and show that fine-tuning learning-to-defer algorithms reduces trip durations. 

\subsection{Experimental Setup}
We consider vehicle deferral for a given driver, $i$, given rain condition, $0 \leq r \leq 1$, and darkness condition, $0 \leq d \leq 1$.
Our dataset consists of $n$ total drivers, each of whom have a mean driving time, $\mu \sim \mathrm{Poisson(35)}$, a rain driving time, $\mu_r \sim \mathrm{Poisson(5)}$, and a darkness driving time, $\mu_d \sim \mathrm{Poisson(5)}$. 
Given $\mu$, $\mu_r$, and $\mu_d$, the amount of time necessary to drive a trip in conditions $(r,d)$ is distributed according to
\begin{equation} 
\mathcal{N}(\mu,\sigma) + r \mathcal{N}(\mu_r,\sigma_r) +  d \mathcal{N}(\mu_d,\sigma_d) 
\end{equation} 

We generate $k$ such trips for each driver; for drivers, $j \neq i$, the points are for training, while for driver $i$, the points are split equally between fine-tuning and testing. 
Autonomous vehicles have three options: they can defer to humans, drive independent of conditions in time $\mathcal{N}(\mu_a,\sigma_a)$, or drive  dependent on conditions in time $\mathcal{N}(\mu_b,\sigma_b) + r \mathcal{N} (\mu_x,\sigma_x) + d \mathcal{N} (\mu_y,\sigma_y)$, where $\mu,\sigma$ are constants chosen to make it possible for the system to reduce average trip time by making accurate defer decisions.
This forces autonomous vehicles to learn both a rejector and classifier function, emulating the joint classification-rejection problem. 
To assist with learning, an additional $l$ unlabeled $(r,d)$ pairs are presented. 
We provide further experimental details in Appendix~\ref{sec:autonomous-appendix}.

\begin{figure*}
    \centering
    \includegraphics[width=250pt]{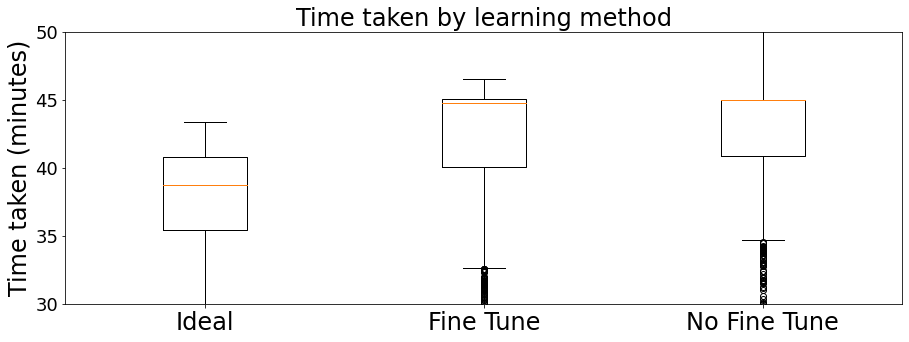}
    \caption{Distribution of time taken with/without fine-tuning and ideally. Each data point represents the average trip duration when fine-tuning on one driver, over $128$ testing trips. Fine-tuning outperforms no fine-tuning by 30 seconds, showing that fine-tuning can assist autonomous vehicles.}
    \label{fig:driving}
\end{figure*}

\subsection{Experimental Results}
To analyze the results, we compute the ideal time, which chooses knowing trip durations beforehand and ignores randomness, and known mean, which chooses based on known $\mu$ values.
Both of these serve as lower bounds on average trip duration.

We find that fine-tuning reduces average trip duration by 30 seconds compared to not fine-tuning, while knowing $\mu$ values can reduce trip durations by 2 minutes 30 seconds. 
Fine-tuning reduces trip duration error by $20\%$ (i.e., it achieves $20\%$ of achievable reduction if $\mu$ values were known), and our results are significant with $p<.001$. 
While the medians between the two learning-to-defer methods are close, the means are wider apart due to large outliers when not fine-tuning.
\section{CIFAR-10 Experiments}
We use the CIFAR-10 image classification task to assess the impact of fine-tuning on human-AI collaboration, using both synthetic human models and real human annotation data. 



\subsection{Experiment Setup}
This task asks users to classify 32x32 images into one of ten non-overlapping categories~[\cite{krizhevsky2009learning}]. 
We consider the effect of fine-tuning when having synthetic or real experts work with AI image classifiers. 

As in~[\cite{mozannar2020consistent}], we create synthetic experts parameterized by $k$, which indicates perfect accuracy on the first $k$ image classes, and random guesses on the last $10-k$. 
Each synthetic expert annotates $500$ images, while real experts annotate $200$ images, which are split $50\%-50\%$ for fine-tuning and testing.
For human expert data, we select 11 annotators from CIFAR-10H~[\cite{peterson2019human}].

We begin by training a human performance function $\hat{h}$ to predict expert success on an image, outputting $0$ if correct, and $1$ otherwise.
We train a WideResNet model~[\cite{zagoruyko2016wide}] on aggregate human data from CIFAR-10H, then fine-tune on data from a single synthetic or real expert.
For self-training, after training $\hat{h}$, we impute predictions on the 50k training labels that lack human predictions, then retrain using high confidence points.
We use the fine-tuned human performance model to learn a combined rejector-classifier, which is an 11-class WideResNet model that outputs 0-9 to predict a class, and 10 to defer.
We evaluate system accuracy over the test images and compare three different approaches: no fine-tuning, baseline fine-tuning, and fine-tuning with self-training.
\subsection{Results}
\begin{figure}[!ht]
    \centering
    \includegraphics[width=150pt]{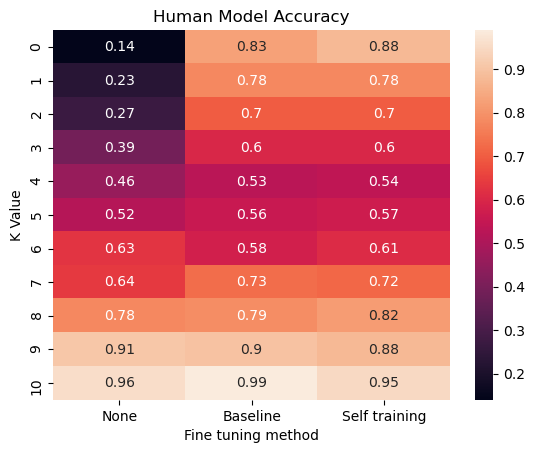}
    \includegraphics[width=150pt]{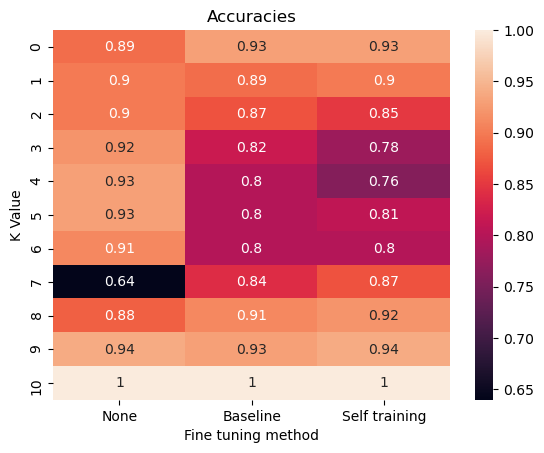}
    \caption{Human model and system accuracy for synthetic annotators. Fine-tuning improves human model accuracy, especially for low values of $k$, though fails to consistently improve system accuracy.}
    \label{fig:human_accuracy}
\end{figure}
Our takeaways from the synthetic and real annotator experiments are:
\begin{enumerate}
    \item \textbf{Fine-tuning can improve expert prediction in some situations}. For synthetic experts, fine-tuning improves human model accuracy, particularly for low values of $k$ (Figure~\ref{fig:human_accuracy}). 
    However, fine-tuning fails to consistently outperform not fine-tuning for real annotators, potentially due to lack of learnability.
    Synthetic experts may pose an easier challenge when compared to real annotators with more nuanced strengths and weaknesses.
    \item \textbf{Correlation between human and system accuracy}. For real experts, when baseline fine-tuning improves human model accuracy, system accuracy also improves $75\%$ of the time.
\begin{figure}[!ht]
    \centering
    \includegraphics[width=150pt]{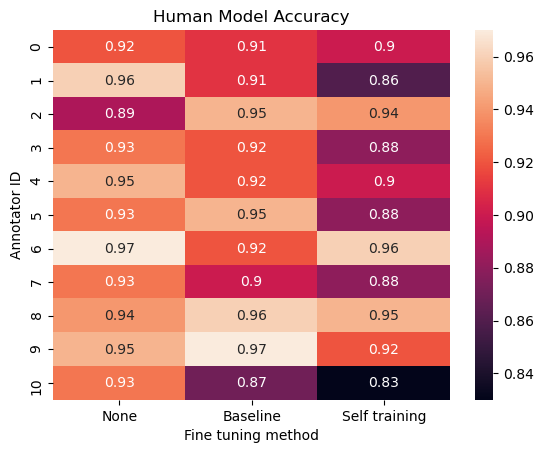}
    \includegraphics[width=150pt]{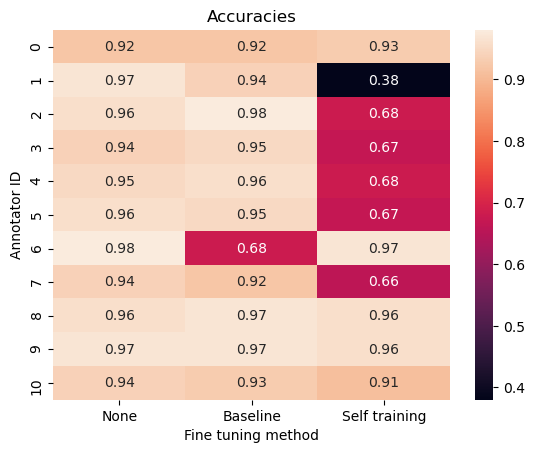}
    \caption{Human model and system accuracy for real annotators. Fine-tuning doesn't consistently improve human model; however, when it helps, system accuracy also tends to improve.}
    \label{fig:system_accuracy}
\end{figure}
\end{enumerate}
Our results indicate that fine-tuning can pick up on simple patterns of skill, but struggles with nuance.
We can potentially improve fine-tuning by exploring more sophisticated semi-supervised learning algorithms, such as graph-based techniques~[\cite{subramanya2014graph}], which can then boost system accuracy.

\section{Conclusion}
We explore the use of individual-specific human-AI collaboration algorithms. 
We find that tailoring learning-to-defer algorithms to individuals through fine-tuning can improve human model and system accuracy in some situations. However, initial fine-tuning approaches struggle to consistently improve human model performance for real experts, though can improve system performance when they do. We propose investigating more advanced semi-supervised techniques to address this.


\clearpage
\bibliographystyle{IEEEtranN}
\bibliography{references}

\begin{thebibliography}{20}
\providecommand{\natexlab}[1]{#1}
\providecommand{\url}[1]{#1}
\csname url@samestyle\endcsname
\providecommand{\newblock}{\relax}
\providecommand{\bibinfo}[2]{#2}
\providecommand{\BIBentrySTDinterwordspacing}{\spaceskip=0pt\relax}
\providecommand{\BIBentryALTinterwordstretchfactor}{4}
\providecommand{\BIBentryALTinterwordspacing}{\spaceskip=\fontdimen2\font plus
\BIBentryALTinterwordstretchfactor\fontdimen3\font minus
  \fontdimen4\font\relax}
\providecommand{\BIBforeignlanguage}[2]{{%
\expandafter\ifx\csname l@#1\endcsname\relax
\typeout{** WARNING: IEEEtranN.bst: No hyphenation pattern has been}%
\typeout{** loaded for the language `#1'. Using the pattern for}%
\typeout{** the default language instead.}%
\else
\language=\csname l@#1\endcsname
\fi
#2}}
\providecommand{\BIBdecl}{\relax}
\BIBdecl

\bibitem[Tran et~al.(2019)Tran, Riveros, and Ravaud]{tran2019patients}
V.-T. Tran, C.~Riveros, and P.~Ravaud, ``Patients’ views of wearable devices
  and ai in healthcare: findings from the compare e-cohort,'' \emph{NPJ digital
  medicine}, vol.~2, no.~1, pp. 1--8, 2019.

\bibitem[Washington(2018)]{washington2018argue}
A.~L. Washington, ``How to argue with an algorithm: Lessons from the
  compas-propublica debate,'' \emph{Colo. Tech. LJ}, vol.~17, p. 131, 2018.

\bibitem[Mozannar and Sontag(2020)]{mozannar2020consistent}
H.~Mozannar and D.~Sontag, ``Consistent estimators for learning to defer to an
  expert,'' in \emph{International Conference on Machine Learning}.\hskip 1em
  plus 0.5em minus 0.4em\relax PMLR, 2020, pp. 7076--7087.

\bibitem[Raghu et~al.(2019)Raghu, Blumer, Corrado, Kleinberg, Obermeyer, and
  Mullainathan]{raghu2019algorithmic}
M.~Raghu, K.~Blumer, G.~Corrado, J.~Kleinberg, Z.~Obermeyer, and
  S.~Mullainathan, ``The algorithmic automation problem: Prediction, triage,
  and human effort,'' \emph{arXiv preprint arXiv:1903.12220}, 2019.

\bibitem[Wilder et~al.(2020)Wilder, Horvitz, and Kamar]{wilder2020learning}
B.~Wilder, E.~Horvitz, and E.~Kamar, ``Learning to complement humans,''
  \emph{arXiv preprint arXiv:2005.00582}, 2020.

\bibitem[Okati et~al.(2021)Okati, De, and
  Gomez-Rodriguez]{okati2021differentiable}
N.~Okati, A.~De, and M.~Gomez-Rodriguez, ``Differentiable learning under
  triage,'' \emph{arXiv preprint arXiv:2103.08902}, 2021.

\bibitem[De et~al.(2021)De, Okati, Zarezade, and
  Rodriguez]{de2021classification}
A.~De, N.~Okati, A.~Zarezade, and M.~G. Rodriguez, ``Classification under human
  assistance,'' in \emph{Proceedings of the AAAI Conference on Artificial
  Intelligence}, vol.~35, no.~7, 2021, pp. 5905--5913.

\bibitem[Krizhevsky et~al.(2009)Krizhevsky, Hinton,
  et~al.]{krizhevsky2009learning}
A.~Krizhevsky, G.~Hinton \emph{et~al.}, ``Learning multiple layers of features
  from tiny images,'' 2009.

\bibitem[Chapelle et~al.(2009)Chapelle, Scholkopf, and Zien]{chapelle2009semi}
O.~Chapelle, B.~Scholkopf, and A.~Zien, ``Semi-supervised learning (chapelle,
  o. et al., eds.; 2006)[book reviews],'' \emph{IEEE Transactions on Neural
  Networks}, vol.~20, no.~3, pp. 542--542, 2009.

\bibitem[Peterson et~al.(2019)Peterson, Battleday, Griffiths, and
  Russakovsky]{peterson2019human}
J.~C. Peterson, R.~M. Battleday, T.~L. Griffiths, and O.~Russakovsky, ``Human
  uncertainty makes classification more robust,'' in \emph{Proceedings of the
  IEEE/CVF International Conference on Computer Vision}, 2019, pp. 9617--9626.

\bibitem[{Zagoruyko} and {Komodakis}(2016)]{zagoruyko2016wide}
S.~{Zagoruyko} and N.~{Komodakis}, ``Wide residual networks,'' in \emph{British
  Machine Vision Conference 2016}, 2016.

\bibitem[Subramanya and Talukdar(2014)]{subramanya2014graph}
A.~Subramanya and P.~P. Talukdar, ``Graph-based semi-supervised learning,''
  \emph{Synthesis Lectures on Artificial Intelligence and Machine Learning},
  vol.~8, no.~4, pp. 1--125, 2014.

\bibitem[Nagar and Malone(2011)]{nagar2011making}
Y.~Nagar and T.~W. Malone, ``Making business predictions by combining human and
  machine intelligence in prediction markets.''\hskip 1em plus 0.5em minus
  0.4em\relax Association for Information Systems, 2011.

\bibitem[Patel et~al.(2019)Patel, Rosenberg, Willcox, Baltaxe, Lyons, Irvin,
  Rajpurkar, Amrhein, Gupta, Halabi, et~al.]{patel2019human}
B.~N. Patel, L.~Rosenberg, G.~Willcox, D.~Baltaxe, M.~Lyons, J.~Irvin,
  P.~Rajpurkar, T.~Amrhein, R.~Gupta, S.~Halabi \emph{et~al.}, ``Human--machine
  partnership with artificial intelligence for chest radiograph diagnosis,''
  \emph{NPJ digital medicine}, vol.~2, no.~1, pp. 1--10, 2019.

\bibitem[Vaccaro and Waldo(2019)]{vaccaro2019effects}
M.~Vaccaro and J.~Waldo, ``The effects of mixing machine learning and human
  judgment,'' \emph{Communications of the ACM}, vol.~62, no.~11, pp. 104--110,
  2019.

\bibitem[Bansal et~al.(2021{\natexlab{a}})Bansal, Wu, Zhou, Fok, Nushi, Kamar,
  Ribeiro, and Weld]{bansal2021does}
G.~Bansal, T.~Wu, J.~Zhou, R.~Fok, B.~Nushi, E.~Kamar, M.~T. Ribeiro, and
  D.~Weld, ``Does the whole exceed its parts? the effect of ai explanations on
  complementary team performance,'' in \emph{Proceedings of the 2021 CHI
  Conference on Human Factors in Computing Systems}, 2021, pp. 1--16.

\bibitem[Bansal et~al.(2019)Bansal, Nushi, Kamar, Lasecki, Weld, and
  Horvitz]{bansal2019beyond}
G.~Bansal, B.~Nushi, E.~Kamar, W.~S. Lasecki, D.~S. Weld, and E.~Horvitz,
  ``Beyond accuracy: The role of mental models in human-ai team performance,''
  in \emph{Proceedings of the AAAI Conference on Human Computation and
  Crowdsourcing}, vol.~7, no.~1, 2019, pp. 2--11.

\bibitem[Madras et~al.(2017)Madras, Pitassi, and Zemel]{madras2017predict}
D.~Madras, T.~Pitassi, and R.~Zemel, ``Predict responsibly: improving fairness
  and accuracy by learning to defer,'' \emph{arXiv preprint arXiv:1711.06664},
  2017.

\bibitem[Keswani et~al.(2021)Keswani, Lease, and
  Kenthapadi]{keswani2021towards}
V.~Keswani, M.~Lease, and K.~Kenthapadi, ``Towards unbiased and accurate
  deferral to multiple experts,'' \emph{arXiv preprint arXiv:2102.13004}, 2021.

\bibitem[Bansal et~al.(2021{\natexlab{b}})Bansal, Nushi, Kamar, Horvitz, and
  Weld]{bansal2021most}
G.~Bansal, B.~Nushi, E.~Kamar, E.~Horvitz, and D.~S. Weld, ``Is the most
  accurate ai the best teammate? optimizing ai for teamwork,'' in
  \emph{Proceedings of the AAAI Conference on Artificial Intelligence},
  vol.~35, no.~13, 2021, pp. 11\,405--11\,414.

\end{thebibliography}


\begin{thebibliography}{10}

\bibitem{bansal2019beyond}
Gagan Bansal, Besmira Nushi, Ece Kamar, Walter~S Lasecki, Daniel~S Weld, and
  Eric Horvitz.
\newblock Beyond accuracy: The role of mental models in human-ai team
  performance.
\newblock In {\em Proceedings of the AAAI Conference on Human Computation and
  Crowdsourcing}, volume~7, pages 2--11, 2019.

\bibitem{bansal2021does}
Gagan Bansal, Tongshuang Wu, Joyce Zhou, Raymond Fok, Besmira Nushi, Ece Kamar,
  Marco~Tulio Ribeiro, and Daniel Weld.
\newblock Does the whole exceed its parts? the effect of ai explanations on
  complementary team performance.
\newblock In {\em Proceedings of the 2021 CHI Conference on Human Factors in
  Computing Systems}, pages 1--16, 2021.

\bibitem{bansal2021most}
Gagan Bansal, Besmira Nushi, Ece Kamar, Eric Horvitz, and Daniel~S Weld.
\newblock Is the most accurate ai the best teammate? optimizing ai for
  teamwork.
\newblock In {\em Proceedings of the AAAI Conference on Artificial
  Intelligence}, volume~35, pages 11405--11414, 2021.

\bibitem{hendrycks2019robustness}
Dan Hendrycks and Thomas Dietterich.
\newblock Benchmarking neural network robustness to common corruptions and
  perturbations.
\newblock {\em Proceedings of the International Conference on Learning
  Representations}, 201.

\bibitem{jeopardy}
Jeopardy.
\newblock Jeopardy j{!}archive.
\newblock \url{https://j-archive.com/}, 2021.
\newblock Accessed: 2021-08-11.

\bibitem{peterson2019human}
Joshua~C Peterson, Ruairidh~M Battleday, Thomas~L Griffiths, and Olga
  Russakovsky.
\newblock Human uncertainty makes classification more robust.
\newblock In {\em Proceedings of the IEEE/CVF International Conference on
  Computer Vision}, pages 9617--9626, 2019.

\bibitem{washington2018argue}
Anne~L Washington.
\newblock How to argue with an algorithm: Lessons from the compas-propublica
  debate.
\newblock {\em Colo. Tech. LJ}, 17:131, 2018.

\bibitem{tran2019patients}
Viet-Thi Tran, Carolina Riveros, and Philippe Ravaud.
\newblock Patients’ views of wearable devices and ai in healthcare: findings
  from the compare e-cohort.
\newblock {\em NPJ digital medicine}, 2(1):1--8, 2019.

\bibitem{mozannar2020consistent}
Hussein Mozannar and David Sontag.
\newblock Consistent estimators for learning to defer to an expert.
\newblock In {\em International Conference on Machine Learning}, pages
  7076--7087. PMLR, 2020.

\bibitem{okati2021differentiable}
Nastaran Okati, Abir De, and Manuel Gomez-Rodriguez.
\newblock Differentiable learning under triage.
\newblock {\em arXiv preprint arXiv:2103.08902}, 2021.

\bibitem{keswani2021towards}
Vijay Keswani, Matthew Lease, and Krishnaram Kenthapadi.
\newblock Towards unbiased and accurate deferral to multiple experts.
\newblock {\em arXiv preprint arXiv:2102.13004}, 2021.

\bibitem{chapelle2009semi}
Olivier Chapelle, Bernhard Scholkopf, and Alexander Zien.
\newblock Semi-supervised learning (chapelle, o. et al., eds.; 2006)[book
  reviews].
\newblock {\em IEEE Transactions on Neural Networks}, 20(3):542--542, 2009.

\bibitem{raghu2019algorithmic}
Maithra Raghu, Katy Blumer, Greg Corrado, Jon Kleinberg, Ziad Obermeyer, and
  Sendhil Mullainathan.
\newblock The algorithmic automation problem: Prediction, triage, and human
  effort.
\newblock {\em arXiv preprint arXiv:1903.12220}, 2019.

\bibitem{madras2017predict}
David Madras, Toniann Pitassi, and Richard Zemel.
\newblock Predict responsibly: improving fairness and accuracy by learning to
  defer.
\newblock {\em arXiv preprint arXiv:1711.06664}, 2017.

\bibitem{subramanya2014graph}
Amarnag Subramanya and Partha~Pratim Talukdar.
\newblock Graph-based semi-supervised learning.
\newblock {\em Synthesis Lectures on Artificial Intelligence and Machine
  Learning}, 8(4):1--125, 2014.

\bibitem{wilder2020learning}
Bryan Wilder, Eric Horvitz, and Ece Kamar.
\newblock Learning to complement humans.
\newblock {\em arXiv preprint arXiv:2005.00582}, 2020.

\bibitem{foret2020sharpness}
Pierre Foret, Ariel Kleiner, Hossein Mobahi, and Behnam Neyshabur.
\newblock Sharpness-aware minimization for efficiently improving
  generalization.
\newblock {\em arXiv preprint arXiv:2010.01412}, 2020.

\bibitem{krizhevsky2009learning}
Alex Krizhevsky, Geoffrey Hinton, et~al.
\newblock Learning multiple layers of features from tiny images.
\newblock 2009.

\bibitem{williams2019rethinking}
Rebecca~A Williams.
\newblock Rethinking deference for algorithmic decision-making.
\newblock 2019.

\bibitem{vaccaro2019effects}
Michelle Vaccaro and Jim Waldo.
\newblock The effects of mixing machine learning and human judgment.
\newblock {\em Communications of the ACM}, 62(11):104--110, 2019.

\bibitem{nagar2011making}
Yiftach Nagar and Thomas~W Malone.
\newblock Making business predictions by combining human and machine
  intelligence in prediction markets.
\newblock Association for Information Systems, 2011.

\bibitem{patel2019human}
Bhavik~N Patel, Louis Rosenberg, Gregg Willcox, David Baltaxe, Mimi Lyons,
  Jeremy Irvin, Pranav Rajpurkar, Timothy Amrhein, Rajan Gupta, Safwan Halabi,
  et~al.
\newblock Human--machine partnership with artificial intelligence for chest
  radiograph diagnosis.
\newblock {\em NPJ digital medicine}, 2(1):1--10, 2019.

\bibitem{de2021classification}
Abir De, Nastaran Okati, Ali Zarezade, and Manuel~Gomez Rodriguez.
\newblock Classification under human assistance.
\newblock In {\em Proceedings of the AAAI Conference on Artificial
  Intelligence}, volume~35, pages 5905--5913, 2021.

\bibitem{zagoruyko2016wide}
Sergey Zagoruyko and Nikos Komodakis.
\newblock Wide residual networks.
\newblock {\em arXiv preprint arXiv:1605.07146}, 2016.

\end{thebibliography}



\appendix 
\section{Related Works}
\textbf{Human-AI Collaboration}. Collaboration between humans and machines has been explored in various domains including finance~[\cite{nagar2011making}] and healthcare~\cite{patel2019human}. 
Collaboration is beneficial as information from both sources can be combined to make decisions~[\cite{nagar2011making}]. 
However, collaboration has risks, as algorithmic predictions can anchor and bias human predictions~\cite{vaccaro2019effects}. 
This problem is further exacerbated when AI solutions are accompanied by explanations~[\cite{bansal2021does}], as humans are more liable to accept explanations regardless of their veracity. 
However, the problem can be mitigated through the development of human mental models, which acknowledge deficits in AI performance~[\cite{bansal2019beyond}]. 
\\ \\
\textbf{Learning to defer}. One area of research within human-AI collaboration is learning to defer, where AI systems defer to humans when unsure. 
For classification problems, the AI system consists of a rejector, which determines whether or not to defer, and a classifier, which solves the task. 
One strategy is to calculate confidence scores for AI and humans, and defer based on whichever is higher~[\cite{raghu2019algorithmic,madras2017predict}]. 
However, these confidence-based approaches ignore the interconnectedness of the rejector and classifier by training the classifier independently. Ideally, the classifier should be trained to focus on tasks that are hard for humans, and to solve this, prior work has developed a joint learning algorithm to simultaneously learn a rejector and classifier through the use of a new loss function~[\cite{wilder2020learning,mozannar2020consistent}]. 
In the case of SVMs, prior work has shown theoretically and experimentally that learning-to-defer algorithms outperform humans and machines separately~[\cite{de2021classification}].
Extensions to the learning-to-defer approach include algorithms for multiple experts~[\cite{keswani2021towards}] and human-initiated deferral~[\cite{bansal2021most}]. 
Our work most directly builds on prior work by Mozannar et al.~[\cite{mozannar2020consistent}], and responds to a call from De et al.~[\cite{de2021classification}] to develop human-AI collaboration algorithms that can be adapted to specific humans. 
\section{Hyperparameters}
For our autonomous vehicle experiments, we use $\sigma=5$, $\sigma_r=\sigma_d=2$, $\mu_a = 45$, $\sigma_a = .001$, $\mu_b = 40$, $\sigma_b = 5$ $\mu_x = \mu_y = 5$, $\sigma_x = \sigma_y = 2$.
Additionally, for both experiments, for baseline fine-tuning, we use $\lambda=2$.


\section{Autonomous Vehicle Details}
\label{sec:autonomous-appendix}
We train a human performance model and a joint rejector-classifier function, which returns $0$ to defer, $1$ to drive independent of conditions, and $2$ to drive dependent on conditions. 
We use a feed-forward neural network to model both the rejector-classifier and the human performance model. 
We compare the baseline with and without fine-tuning to develop the human performance model, and use the model to impute predictions for unlabeled data. 
For our experiments, we use $n=10$, $k=256$, and $l=1000$.

\end{document}